\documentclass[a4paper]{article}
\usepackage{placeins}
\usepackage{INTERSPEECH2016}
\usepackage{textcomp}
\usepackage{hyperref}
\usepackage{url}
\usepackage{booktabs} 
\usepackage{graphicx,subfigure}
\usepackage{color}
\usepackage{float}
\usepackage{amsmath,amsfonts,amssymb,bm}
\usepackage[font=small,labelfont=bf]{caption}
\usepackage[english]{babel}
\usepackage[T1]{fontenc}
\usepackage[ansinew]{inputenc}
\usepackage{filecontents}

\def\vec#1{\ensuremath{\bm{{#1}}}}

\ninept

\renewcommand{\vec}[1]{\boldsymbol{#1}}

\newcommand{\mat}[1]{\mathbf{#1}}

\usepackage{adjustbox}
\usepackage{array}
\usepackage{booktabs}

\newcolumntype{R}[2]{%
    >{\adjustbox{angle=#1,lap=\width-(#2)}\bgroup}%
    l%
    <{\egroup}%
}

\newcommand{\word}[1]{``\emph{#1}''} 

\newcommand{\tok}[1]{\vec{x}_{#1}}

\newcommand{\stok}[1]{\vec{s}_{#1}}
\newcommand{\red}[1]{\emph{\color{blue}#1}}
\newcommand{\hidden}[1]{\vec{h}_{#1}}
\newcommand{\out}[1]{\vec{y}_{#1}}

\newcommand{\func}[2]{\vec{f}_{#1}\left( #2 \right)}
\newcommand{\outfunc}[2]{g_{#1}\left( #2 \right)}
\newcommand{\wmat}[1]{\mat{W}_{#1}}
\newcommand{\oconv}{{\sc R-Conv}}
\newcommand{\ldaconv}{{\sc LDA-Conv}}
\newcommand{\rldaconv}{{\sc R-LDA-Conv}}
\newcommand{\poster}{{\sc Poster}}
\newcommand{\responder}{{\sc Responder}}
\newcommand{\etal}{\emph{et al.}~}

\newcommand{\softmax}[1]{\text{softmax}(#1)} 

\title{LSTM based Conversation Models}
\makeatletter
\def\name#1{\gdef\@name{#1\\}}
\makeatother \name{{\em Yi Luan$^1$, Yangfeng Ji$^2$, Mari Ostendorf$^1$}}

\address{$^1$Department of Electrical Engineering, University of Washington, Seattle, WA 98195 \\
  $^2$School of Interactive Computing, Georgia Institute of Technology, Atlanta, GA 30332 \\
  {\small \tt luanyi@uw.edu, jiyfeng@gatech.edu, ostendor@uw.edu}
}

\begin{document}
\maketitle

\begin{abstract}
In this paper, we present a conversational model that incorporates both context and participant role for two-party conversations. Different architectures are explored for integrating participant role and context information into a Long Short-term Memory (LSTM) language model. The conversational model can function as a language model or a  language generation model. Experiments on the Ubuntu Dialog Corpus show that our model can capture multiple turn interaction between participants. The proposed method outperforms a traditional LSTM model  as measured by language model perplexity and response ranking.  Generated responses show characteristic differences between the two participant roles.
\end{abstract}



\section{Introduction}
\label{sec:intro}

As automatic language understanding and generation technology improves, there is increasing interest in building human-computer conversational systems, which can be used for a variety of applications such as travel planning, tutorial systems or chat-based technical support.  Most work has emphasized understanding or generating a word sequence associated with a single sentence or speaker turn, potentially leveraging the previous turn.  Beyond local context, language use in a goal-oriented conversation reflects the global topic of discussion, as well as the respective role of each participant. In this work, we introduce a conversational language model that incorporates both local and global context together with participant role. 

In particular,
participant roles (\emph{or} speaker roles) impact content of a sentence in terms of both the information to be communicated and the interaction strategy, affecting both meaning and conversational structure. For example, in broadcast news, speaker roles are shown to be informative for discovering the story structures \cite{barzilay2000rules}; they impact speaking time and turn-taking \cite{Hutchinson+10}; and they are associated with particular phrase patterns \cite{Zhang+13}.  In online discussions, speaker role is useful for detecting authority claims \cite{marin2011detecting}. Other work shows that in casual conversations, speakers with different roles are likely to use different discourse markers~\cite{fuller2003influence}. 
For the Ubuntu technical support data used in this study,
Table~\ref{tab:word} illustrates differences in the distributions of frequent words for the poster vs.\ responder roles. The \poster~ role tends to raise questions using words {\it anyone}, {\it how}. The \responder~ role tends to use directive words ({\it you}, {\it you're}), hedges ({\it may}, {\it might}) and words related to problem solving ({\it sudo}, {\it check}).

 \begin{table*}[t]
  \centering
 {\footnotesize
   \begin{tabular}{lp{11cm}}
    \toprule
    $p(w|\poster)/p(w|\responder)$ & hi, hello, anyone, hey, guys, ideas, thanks, thank, my, how, am, ??, cannot, I'm, says\\
    \midrule
    $p(w|\responder)/p(w|\poster)$ & you're, your, probably, you, may, might, sudo, ->, search, sure, ask, maybe, most, check, try\\
    \bottomrule
  \end{tabular}}
  \caption{\small{Top 15 words based on the role likelihood ratio out of the subset with word count > 6k.}}
  \label{tab:word}
\end{table*}


Specifically, we propose a neural network model that builds on recent work in response generation, integrating different methods that have been used for capturing local (previous sentence) context and more global context, and extending the network architecture to incorporate role information. The model can be used as a language model, as in speech recognition or translation, but our focus here is on response generation.  Experiments are conducted with Ubuntu chat logs, using language model perplexity and response ranking, as well as qualitative analysis.





\section{Related Work}
\label{sec:related}



Data-driven methods are now widely used for building conversation systems. With the popularity of social media, such as Twitter, Sina Weibo, and online discussion forums, it is easier to collect conversation text~\cite{wang2013dataset,lowe2015ubuntu}. Several different data-driven models have been proposed to build conversation systems. Ritter \etal \cite{ritter2011data} present a statistical machine translation based conversation system. Recently, neural network models have been explored. The flexibility of neural network models opens the possibility of integrating different kinds of information into the generation procedure. For example, Sordoni \etal \cite{sordoni2015neural} present a way to integrate contextual information via feed-forward neural networks. Li \etal propose using Maximum Mutual Information (MMI) as the objective function in neural models in order to produce more diverse and interesting responses. Shang \etal \cite{shang2015neural} introduce the attention mechanism into an encoder-decoder network for a conversation model. Most similar to our work is the Semantic Controlled LSTM (SC-LSTM) proposed by Wan \etal \cite{wen2015semantically}, where a Dialog-act component is introduced into the LSTM cell to guide the generated content. In this work, we utilize the role information to bias response generation without modifying LSTM cells. 

Efficiently capturing local and global context remains a open problem in language modeling. Different ways of modeling document-level context has been explored in \cite{ji2015document} and \cite{lin2015hierarchical} based on the LSTM framework.  Luan \etal \cite{luan2015efficient} proposed a multi-scale recurrent architecture to incorporate both word and turn level context for spoken language understanding tasks. In this paper, we use a similar approach as \cite{mikolov2012context}, explicitly using Latent Dirichlet Analysis (LDA) as global-context feature to feed into RNNLM.

Early work on incorporating local context in conversational language modeling is described in \cite{JiBilmes04} conditioned on the most recent word spoken by other speakers.  Hutchinson \etal \cite{hutchinson2013rank,hutchinson2015sparse} improve log-bilinear language model by introducing a multi-factor sparse matrix that could capture speaker role and topic information. In addition, Huang \etal \cite{huang2008modeling} show that language models with role information significantly reduce word error rate in speech recognition. Our work differs from these approaches in using an LSTM. Recently, Li \etal propose using an additional vector to LSTM in order to capture personal characteristics of a speaker \cite{li2016persona}. In this work, we utilize both a global topic vector and role information, where a role-specific weight matrix biases the word distributions for different roles.



\section{Model}
\label{sec:model}

In this section, we propose an LSTM based framework that integrating participant role and global topic of the conversation. As discussed in section~\ref{sec:intro}, the assumption is, given the same context, each role has its own preference of picking words to generate a response. Each generated response should be both topically related to the current conversation and coherent with the local context. 
\subsection{Recurrent Neural Network Language Models}
We start building a response generation model~\cite{sordoni2015neural} by using a recurrent neural network language model (RNNLM)~\cite{mikolov2010recurrent}. In general, a RNNLM is a generative model of sentences. For a sentence consisted of word sequence $x_{1},\dots,x_{I}$, the probability of $x_{i}$ given $x_{1},\dots,x_{i-1} \triangleq \tok{\leq i-1}$ is 
\begin{equation}
  \label{eq:rnnlm}
  p(x_{i}|\tok{\leq i-1}) \propto \outfunc{\tau}{\hidden{i}}
\end{equation}
\noindent where $\hidden{i}$ is the current hidden state and $\outfunc{\tau}{\cdot}$ is the probability function parameterized by $\tau$:
\begin{equation}
  \label{eq:our-imp}
  \outfunc{\tau}{\hidden{i}}=\softmax{\wmat{\tau}\hidden{i}}, 
\end{equation}

\noindent where $\wmat{\tau}$ is the output layer parameter. The hidden state $\hidden{i}$ is computed recurrently as 
\begin{equation}
  \label{eq:hidden}
  \hidden{i}=\func{\theta}{x_{i},\hidden{i-1}}. 
\end{equation}
$\func{\theta}{\cdot}$ is a nonlinear function parameterized by $\theta$. We use an LSTM~\cite{hochreiter1997long} since it is good at capturing long-term dependency, which is an objective for our conversation model. 

\subsection{Conversation Models with Speaker Roles}


To build a conversation model with different participant roles, we extend a RNNLM in two respects. First, to capture the variability from different participant roles, we incorporate role-based information into the generation procedure. Second, to model a conversation instead of single turns, our model adjoins RNNLMs for all turns in sequence to model the whole conversation. 

More specifically, consider two adjacent turns\footnote{In our formulation, we use one turn as the minimal unit as multiple sentences in one turn share the same role.} $\tok{t-1}=\{x_{t-1,i}\}_{i=1}^{N_{t-1}}$ and $\tok{t}=\{x_{t,i}\}_{i=1}^{N_t}$ with their participant role $r_{t-1}$ and $r_{t}$ respectively. $N_t$ is the number of words in the $t$-th turn. To build a single model for the entire conversation, we simply concatenate the RNNLMs for all sentences in order. Concatenation changes the way of computing the first hidden state in each utterance (except the first utterance in the conversation). Considering the two turns $\tok{t-1}$ and $\tok{t}$, after concatenation, the computation of the first hidden state in turn $\tok{t}$, $\hidden{t,1}$, is
\begin{equation}
  \label{eq:hidden-t-1}
  \hidden{t,1} = \func{\theta}{x_{t,1},\hidden{t-1,N_{t-1}}}.
\end{equation}
As we will see from section~\ref{sec:exp}, this simple solution can capture the long-term contextual information.

We introduce the role-based information by defining a role-dependent function $\outfunc{\tau,r}{\cdot}$. For example, the probability of $x_{t,i}$ given $x_{t,i-1},\dots,x_{1,1}\triangleq \tok{\leq t, \leq i-1}$ and its role $r_t$ is
\begin{equation}
  \label{eq:prob-role}
  p(x_{t,i}|\tok{\leq t, \leq i-1}, r_t) \propto \outfunc{\tau,r_t}{\hidden{t,i}}.
\end{equation}
where the $\outfunc{\tau, r_t}{\cdot}$ is also parameterized by role $r_t$. In our implementation, we use 
\begin{equation}
  \label{eq:our-imp}
  \outfunc{\tau,r_t}{\hidden{t,i}}=\softmax{\wmat{\tau}(\wmat{r_t}\hidden{t,i})}, 
\end{equation}
where $\wmat{\tau}\in\mathbb{R}^{V\times H}$, $\wmat{r_t}\in\mathbb{R}^{H\times H}$, $V,H$ are the vocabulary size and hidden layer dimension respectively. Even $\wmat{\tau}$ is shared across the entire conversation model, $\wmat{r_t}$ is role-specific. This linear transformation defined in Eq.~\ref{eq:our-imp} is easy to train in practice and appears to capture role information. This model is named the \oconv~ model, as the role-based information is introduced in the output layer. 

Despite the difference between the two models, they can be learned in the same way, which is similar to training a RNNLM~\cite{mikolov2010recurrent}. Following the way of training a language model, the parameters could be learned by maximizing the following objective function 
\begin{equation}
  \sum_{k}\sum_{t}\ell(x_{t,k+1}, \out{t,k})
\end{equation}
where $\out{t,k}$ is the prediction of $x_{t,i+1}$.  $\ell(\cdot,\cdot)$ can be any loss function for classification task. We choose cross entropy~\cite{murphy2012machine} as the loss function $\ell(\cdot,\cdot)$, because it is a popular objective function used in training neural language models. 

As a final comment, if we eliminate the role information, \rldaconv~ will be reduced to an RNNLM. To demonstrate the utility of role-based information, we will use an RNNLM over conversations as a baseline model. 

\subsection{Incorporating global topic context}
In order to capture long-span context of the conversation, inspired by  \cite{mikolov2012context}, we explicitly include a topic vector representing all previous dialog turns. 
We use Latent Dirichlet Allocation (LDA) to achieve a compact vector-space representation.
This procedure maps a bag-of-words representation of a document into a low-dimensional vector which is conventionally interpreted as a topic representation. For each turn $\tok{t}$, we compute the LDA representation for all previous turns 
\begin{equation}
\stok{t}=\func{LDA}{\tok{1}, \tok{2},\dots,\tok{t-1}}
\end{equation}
where $\func{LDA}{\cdot}$ is the LDA inference function as in \cite{blei2003latent}. Then $\stok{t}$ is concatenated with hidden layer $\hidden{t,i}$ to predict $\tok{t,i}$.
\begin{equation}
  \label{eq:prob-role}
  p(x_{t,i}|\tok{\leq t, \leq i-1}) \propto \outfunc{\tau}{[\hidden{t,i}^\top \quad \stok{t}^\top]^\top}.
\end{equation}
This model is named \ldaconv. We assume that by including $\stok{t}$ into output layer, the predicted word would be more topically related with the previous turns, thus allowing the recurrent part to learn more local context information.

\begin{figure}[tb]
\centering
\includegraphics[width=6.5cm]{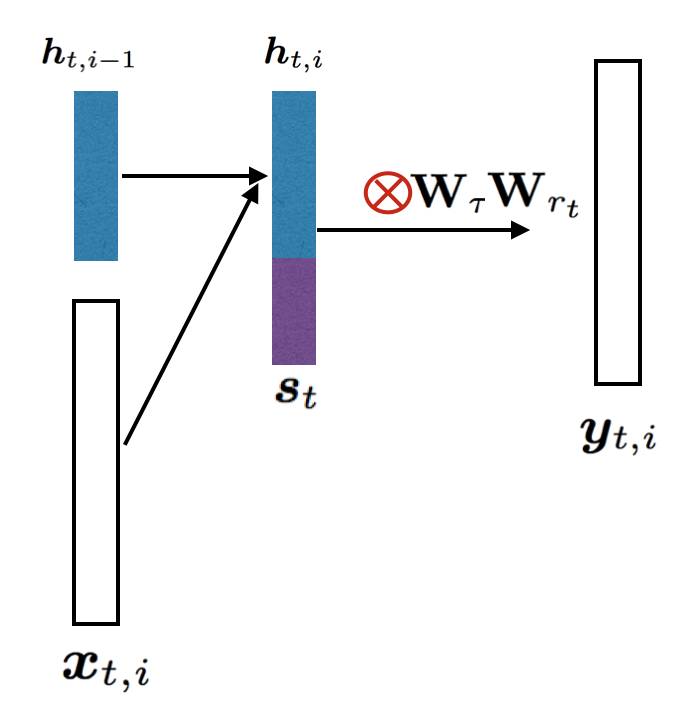}                      
\caption{\small{The \rldaconv~ model. The turn-level LDA feature $\stok{t}$ is concatenated with word-level hidden layer $\hidden{t,i}$ and the output weight matrix $\wmat{r_t}$ is role specific.}}
\label{fig:corr}
\end{figure}

When incorporating both the global topic vector and the role factor, the conditional probability of $\tok{t,i}$ is
\begin{equation}
  \label{eq:prob-role}
  p(x_{t,i}|\tok{\leq t, \leq i-1}, r_t) \propto \outfunc{\tau,r_t}{[\hidden{t,i}^\top \quad \stok{t}^\top]^\top}.
\end{equation}
We call this model, illustrated in Figure \ref{fig:corr}, \rldaconv. 

\section{Experiments}
\label{sec:exp}
We evaluate our model from different aspects on the Ubuntu Dialogue Corpus \cite{lowe2015ubuntu}, which provides one million two-person conversations extracted from Ubuntu chat logs. The conversations are about getting technical support for various Ubuntu-related problems. In this corpus, each conversation contains two users with different roles, {\bf \poster}: the user in this conversation who initializes the conversation by proposing a technical problem; {\bf \responder}: the other user who tries to provide technical support. For a conversation, we replace the user of each turn with the corresponding role.

\subsection{Experimental setup}

Our models are trained in a subset of the Ubuntu Dialogue Corpus, in which each conversation contains 6 - 20 turns. The resulting data contains 216K conversations in the training set, 10k conversations in the test set and 13k conversations in the development set. We use a Twitter tokenizer~\cite{owoputi2013improved} to parse all utterances in the conversations. The vocabulary is constructed on the training set with filtering out low-frequency tokens and replacing them with \word{UNKNOWN}. The vocabulary size is fixed to include 20K most frequent words. We did not filter out emoticons, instead we treat them as single tokens.

The LDA model is trained using all conversations in training data, where each conversation is treated as an individual training instance. We use \textit{Gensim} \cite{rehurek_lrec} for both training and inference. There are three hyper-parameters in our models: the dimension of word representation $K$, the hidden dimension $H$ and the number of topics $M$ in LDA model. We use grid search over $K, H\in\{16, 32, 64, 128, 256\}$, $M\in\{50, 100\}$, and select the best combination for each model using the perplexity on the development set. We use stochastic gradient descent with the initial learning rate $\lambda=0.1$ to train all the models. 

\subsection{Evaluation Metrics}

Evaluation on response generation is an emerging research field in data-driven conversation modeling. Due to the variety of possible responses for a given context, it is too conservative to only compare the generated response with the ground truth. For a reasonable evaluation, the $n$-gram based evaluation metrics including {\sc Bleu}~\cite{papineni2002bleu} and $\Delta${\sc Bleu}~\cite{galley2015deltableu} require multiple references for one given context. One the other hand, there are indirect evaluation methods, for example, ranking based evaluation~\cite{lowe2015ubuntu,shang2015neural} or qualitative analysis~\cite{vinyals2015neural}. In this paper, we use both ranking-based evaluation (Recall@$K$ ~\cite{manning2008introduction}) across all models, and leave the $n$-gram based evaluation for future work. To compute the Recall@K metric of one model given $K$, the model is used to select the top-$K$ candidates, and it is counted as correct if the ground-truth response is included. In addition to Recall@K, we also evaluate the different models based on test set perplexity.

To understand the chat conversations requires intensive knowledge of Ubuntu even for human readers. Therefore, the qualitative analysis focuses mainly on the capacity of capturing role information, not the justification of responses as valid answers to the technical questions.

\begin{table}
  \centering
{\footnotesize
  \begin{tabular}{lllll}
    \toprule
    Model  & $K$ & $H$ & $M$ & Perplexity\\
    \midrule
    Baseline & 32 & 128 & - & 54.93\\
    \oconv & 256 & 128 & - &48.89 \\
    \ldaconv & 256 & 128 & 100 &51.13 \\
    \rldaconv & 256 & 128 & 50 &\textbf{46.75} \\
    \bottomrule
  \end{tabular}}
  \caption{The best perplexity numbers of the three models on the development set.}
  \label{tab:ppl}
\end{table}

\begin{table}
  \centering
 {\footnotesize
   \begin{tabular}{lllll}
    \toprule
    Metric & Baseline & \oconv & \ldaconv & \rldaconv \\
    \midrule
    Recall@1 & 0.12 &  0.15 & 0.13 & \textbf{0.16}\\
    Recall@2 & 0.22 & 0.25 & 0.24 & \textbf{0.26}\\
    \bottomrule
  \end{tabular}}
  \caption{The performance of response ranking with Recall@$K$.}
  \label{tab:select}
\end{table}

\subsection{Quantative Evaluation}

Experiments in this section compare the performance of \ldaconv, \oconv~ and \rldaconv~ to the baseline LSTM system.

\subsubsection{Perplexity}
The best perplexity numbers from the three models are shown in Table~\ref{tab:ppl}. \rldaconv~ gives the lowest perplexity among the four models, nearly 8 points improvement over the baseline model.  
Comparing role vs.\ global topic, role has a bigger improvement on perplexity of $11\%$ reduction for role vs.\ $7\%$ for LDA topic. Combining both leads to a $15\%$ reduction in perplexity. 
To simplify the comparison, in the following experiments, we only use the best configuration for each model. 

\subsubsection{Response ranking}
The task is to rank the ground-truth response with some randomly-selected sentences for a given context. For each test sample, we use the previous $t-1$ sentences as context, trying to select the best $t$th sentence. We randomly select 9 turns from other conversations in the dataset, replacing their role with the ground truth label. As we noticed that sentences from the background channel, like \word{yes}, \word{thank you}, could fit almost all the conversations with various context. To distinguish the background channel from some contentful sentences, we sample the negative examples with the ground-truth sentence length as a constraint --- samples with the similar length ($\pm$ 2 words) are selected as negative examples. 

The Recall@$K$ are shown in Table~\ref{tab:select}. Both \oconv~ and \ldaconv~ are better than baseline result, while \rldaconv~ gives the best performance overall. Both role factors and topic feature are acting positively in ranking ground-truth responses. Even though no role information is explicitly used in the baseline model, the contextual information itself could be a useful hint to rank the ground-truth response higher. Therefore, the performance of the baseline model is still better than random guess.  Again, role has a bigger effect than topic, and the combination gives the best results, but differences in Recall@$K$ performance are small.


\subsection{Qualitative Analysis}

For qualitative analysis, the best \rldaconv\ model is used to generate role-specific responses, and we examined a number of examples to determine whether the generated response fit into the expected speaker role. We include two examples in Table~\ref{tab:generation1} and Table~\ref{tab:generation2} due to the page limitations. For each case, we have responses generated for each of the possible roles: a further question for the \poster\ and a potential solution for the \responder.

As we can see from the context part of Table~\ref{tab:generation1}, different roles clearly have different behaviors during the conversation.  Ignoring the validity of this potential solution, this generated response is consistent with our expectation of the \responder~ role. The response of \poster~ seems quite plausible. The reply of \responder~  is clearly the right style but more domain information in the topic vector could lead to a more useful solution.

Table~\ref{tab:generation2} shows another example to demonstrate the difference between the \poster~ and \responder~ roles. In this example, the response for the \responder~ is not a potential solution but a question to the \poster. Unlike the generated question for the \poster~ role in the previous example, the purpose of \responder 's question is to ask some further details in order to provide a simpler solution. The \poster 's  response also fits well in the local context as well as global topic of ubuntu installation, claiming the difficulty of implementing the \poster 's suggestion. At the same time, the generated responses also show the necessity of incorporating certain domain knowledge into a domain-specific conversation system, which will be explored in future work.



\section{Summary}
We propose an LSTM-based conversation model by incorporating role factor and topic feature to model different word distribution for different roles. We present three models: \oconv, \ldaconv\ and \rldaconv, by incorporating role factors and topic features into output layer. We evaluate the model using both perplexity and response ranking. Both \oconv\ and \ldaconv\ outperform the baseline model on all tasks. The model \rldaconv\ gives the best performance by combining the two components. In addition, the generation results demonstrate the topical coherence and differences in responses associated with different roles. Besides role and topic, our model structure can be generalized to include more supervised information. For future work, we would incorporate supervised domain knowledge into our model to improve the topic relevance of the response.

\FloatBarrier
\begin{table}[!htbp]
  \centering
  {\footnotesize
  \begin{tabular}{rp{5.5cm}}
    \toprule
    Role & Utterance\\ 
    \midrule
    \red{Context} & \\ [0.2em]
	\poster & hey people . i have a disk from someone who accidentally overwrote his ext3 partition with a ntfs partition ...... any tips on how to recover files from the erased ext3 partition ? the ntfs partion was created using `` fast `` formatting ( i.e. only the partition table was erased )\\ [0.2em]
    \responder & you can rebuild a partition table .. brb\\ [0.2em]
    \poster & is that possible ? how ? and he also started writing to the ntfs partition , so at least some data will be lost , hopefully some is still recoverable\\ [0.2em]
    \responder & at that point , he might as well write it off or pay a professional to do it\\ [0.2em]
    \poster & ere4si : is there some guide somewhere on how to make a live cd using the minimal iso ? it doesn't even have a squashfs root filesystem\\ [0.2em]
    \midrule
    \red{Generation}& \\ [0.2em]
    \poster & can you p me to that ? \\ [0.2em]
    \responder & `` sudo fdisk -l `` then use the fstab of the new permissions . then use that combination of * to recover backup\\ [0.2em]
    \midrule
    \red{Ground truth} & \\ [0.2em]
    \responder & make an iso on usb bootable\\
    \bottomrule
  \end{tabular}
  }
  \caption{Response generation example (providing solution)}
  \label{tab:generation1}
\end{table}

\FloatBarrier
\begin{table}[!htbp]
  \centering
  {\footnotesize
  \begin{tabular}{rp{5.5cm}}
    \toprule
    Role & Utterance\\ 
    \midrule
    \red{Context} &\\ [0.2em]
    \poster & question : i am currently installing ubuntu . in the " prepare partitions " dialog box , should i check both ext3 ' /' and ext3 ' /home ' to be formatted ? \\ [0.2em]
     \responder & did you have a previous install of ubuntu on there\\ [0.2em]
    \poster & no\\ [0.2em]
    \responder & err kazol\_ not mount , write to it i mean \\ [0.2em]
    \responder &  then you can format them both\\ [0.2em]
    \midrule
    \red{Generation} &\\ [0.2em]
     \poster & i know , i don't know how to do this . bad , even if i try an encrypted install of ubuntu ... this means roller failed to mount it , so not the default .\\ [0.2em]
    \responder & or something similar . are you trying to eject net crapped on there ? \\ [0.2em]
    \midrule
    \red{Ground truth} &\\ [0.2em]
    \responder & ok thanks\\
    \bottomrule
  \end{tabular}
  }
  \caption{Response generation example (clarification of the problem)}
  \label{tab:generation2}
\end{table}

\newpage
\eightpt
\bibliographystyle{IEEEtran}
\bibliography{ref}
\end{document}